# Gradient Based Seeded Region Grow method for CT Angiographic Image Segmentation

[1]Harikrishna Rai G.N, [2] T.R.Gopalakrishnan Nair,
[1]Research Associate, RIIC, D S Institutions, SET
labs Infosys, Bangalore, hk_rai@yahoo.com
[2]Director –Research and Industry, Senior Member IEEE, trgnair@ieee.org
Dayananda Sagar Institutions, Bangalore

**Abstract – Segmentation of medical images using seeded region growing technique is increasingly becoming a popular method because of its ability to involve high-level knowledge of anatomical structures in seed selection process. Region based segmentation of medical images are widely used in varied clinical applications like visualization, bone detection, tumor detection and unsupervised image retrieval in clinical databases. As medical images are mostly fuzzy in nature, segmenting regions based intensity is the most challenging task. In this paper, we discuss about popular seeded region grow methodology used for segmenting anatomical structures in CT Angiography images. We have proposed a gradient based homogeneity criteria to control the region grow process while segmenting CTA images.**

## 1. INTRODUCTION

Medical image segmentation is one of the most challenging problems in healthcare industry and has been studied extensively in the last few decades. One of the common behaviors of medical images is that they are inherently fuzzy [1] in most of the cases and do not exhibit discrete boundaries posing major challenge for clear segmentation of desired structure within the image. As medical imaging technology has grown tremendously, there are a number of modalities available and terabytes of images are generated everyday in healthcare environment. Difference in these images due to different body parts scanned through varied modalities for different pathological needs make the development of intelligent and efficient unsupervised image segmentation techniques necessary as well as challenging. Segmentation of medical images in 2D or 3D domain has several advantages and applications in healthcare industry. Key application areas include visualization, volume estimation of objects of interest, detection of abnormalities like tumors, tissue quantification, classification and elimination of unnecessary structures in the areas of interest within an image.

In healthcare, Computed Tomographic Angiography (CTA) scans are widely used for artery and vein visualization especially in Brain and heart. This technology allows the detection of cerebral aneurysms, arterial stenosis, sediments in vessels, narrowed vessels and other vascular brain anomalies. CTA studies are CT scans where a contrast agent is applied via intravenous injection. Often the contrast-enhanced scan is accompanied by a native scan to be able to support vessel visualization by subtraction techniques [2]. Also modern CTA scanners are very fast and can produce up to 1500 slices per minute. Hence manual segmentation of these huge volumes of images is impossible and hence calls for a fast and robust segmentation with minimum user assistance.

In general imaging science, Segmentation is an important step in any image analysis process where we take image as input and the output is some detailed description of the scene or object. Figure 1 depicts the steps involved in typical image analysis workflow showing segmentation as key step for succeeding image representation and recognition stages. Segmentation technique basically divides the spatial domain pixels into meaningful parts or regions. This meaningful part created can be a complete object or a part of it. The segmentation process makes use of some physically measured image features and its performance is measured based on the meaning associated with the extracted regions. So segmentation is considered as a psycho-physical challenge [3]. Also, the extracted region must have some meaning with respect to the given image, making it difficult to model it mathematically. Hence segmentation is the most challenging problem where there is no generic solution which can work well for all kind of images. This makes users to make certain assumptions and develop segmentation algorithm which is specific to problem in hand.

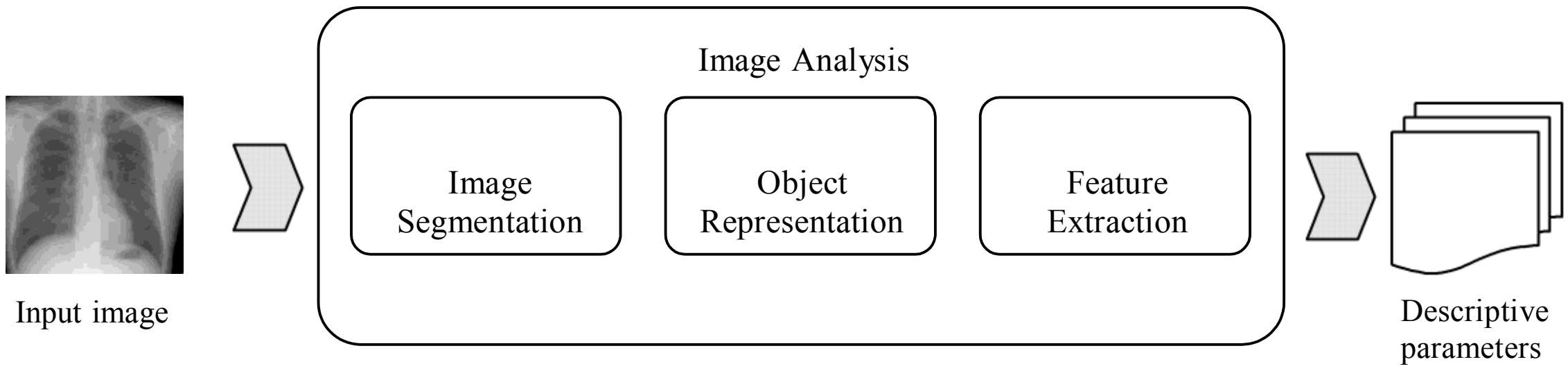

Figure 1- Steps in Image analysis

At high level, image segmentation techniques can be categorized into five approaches [4] thresholding based methods [5, 6], boundary-based methods, region-based methods, hybrid techniques and active contour model based methods. Simple thresholding based methods relies on the brightness constant called threshold value and segment the pixels in the original image according to chosen threshold value. Thresholding method works well in images which has a bimodal distribution. Therefore algorithm can perform well on simple images with bimodal intensity distribution but fails on medical images which do not have bimodal distribution of intensity. In this case, thresholding result cannot partition the images into various anatomical structures correctly. Also this kind of segmentation neglects all spatial information of the image and is quite sensitive to noise.

Boundary-based methods [7] rely on the pixel characteristic which changes rapidly at the boundary between two regions. In this process, in the initial phase, edge detector operators are used for detecting the edge pixels and later these edges are modified to produce the close curves representing the boundaries between adjacent regions. Disadvantage of this method is the difficulty in converting the edge pixels into the close boundary in medical images. Region based segmentation methods are complementary to boundary-based method, this method uses a fundamental concept that neighboring pixels within the one region have similar characteristics. Active contour or Snack techniques [8] are a special category of segmentation, they mainly work based on energy minimization principles. This kind of method is suitable for finding regions with high frequency variation with dominant edges whose gray scale intensities are significantly different from the surrounding region into the images. The user needs to interact with the segmentation process by placing an initial contour around the expected region and active contour techniques will deform these contours to fit the exact region boundaries. Hybrid segmentation methods work mainly based on the combination of fundamental segmentation methods and some additional knowledge based methods.

Segmentation process is also classified into different methods based on the user interaction level needed in the segmentation process. They are - manual segmentation, semi-automatic segmentation and automatic segmentation [12]. Each of these approaches have their own pros and cons. Manual segmentation by domain experts is the most accurate but time-consuming. Semi-automatic segmentation requires the minimal user intervention to provide a small amount of input and parameter to facilitate accurate segmentation. Automatic segmentation does not require any user input and, thus, is much more difficult to obtain accurate results. In applications which involves processing large number of images, automatic segmentation is preferred but at the cost of accuracy.

## 2. SEGMENTATION PROPERTIES

Basic segmentation properties as explained in [3] let ‘I’ be the spatial domain definition of the given image. Then image segmentation technique divides image ‘I’ in to ‘n’ non-overlapping regions represented by $R_i$ *(i = 1, 2, 3,….n)* such that

a) $\bigcup_{i=1}^{n} R_i = I$

b) $R_i \cap R_j = \phi$
c) H ( $R_i$ ) = TRUE
d) H( $R_i$ Ù $R_j$ ) = FALSE if $R_i$ and $R_j$ are adjacent

Where *H(R)* represents the homogeneity criterion, which is based on the feature values established for the segmentation purpose over the region *R*. Also the regions $R_i$ and $R_j$ are considered to be adjacent if a pixel belonging to $R_i$ is a neighbour of some pixels of $R_j$ and vice versa. Property (a) ensures that every pixel in the image is ensured to belong to any one of the non overlapping sub-regions. Second property (b) guarantees that one pixel belongs to only one region in a given image. Third property (c) ensures that the region satisfies the homogeneity criteria defined by the user. Finally the fourth property ensures that the maximality of each region is satisfied.

## 3. SEEDED REGION GROWING METHOD

Seeded region growing (SRG) method for segmentation introduced by [9], is a simple and robust method of segmentation which is rapid and free of tuning parameters. User control over the high level knowledge of image components in the seed selection process makes it a better choice for easy implementation and applying it on a larger dataset. The only drawback of SRG algorithm is the difficulty in automating seed generation and dependency of output on order sorting of pixel as different order of processing pixels during region grow process leads to different final segmentation results.

Seeded region growing approach to image segmentation is to segment an image into regions with respect to a set of *q* seeds as presented in [10] is discussed here. Given the set of seeds, *S1,S2, . . .,Sq,* each step of SRG involves identifying one additional pixel to one of the seed sets. Moreover, these initial seeds are further replaced by the centroids of these generated homogeneous regions, *R1,R2, . . .,Rq,* by involving the additional pixels step by step. The pixels in the same region are labeled by the same symbol and the pixels in variant regions are labeled by different symbols. All these labeled pixels are called the allocated pixels, and the others are called the unallocated pixels.

Let *H* be the set of all unallocated pixels which are adjacent to at least one of the labeled regions.

$$H = \left\{ (x,y) \notin \bigcup_{i=1}^{q} R_i \,|\, N(x,y) \cap \bigcup_{i=1}^{q} R_i \neq \emptyset \right\}$$

where *N(x,y)* is the second-order neighborhood of the pixel *(x,y)*.
For the unlabeled pixel *(x,y) ε H, N(x,y)* meets just one of the labeled image region *Ri* and define *Ψ(x,y) ε {1, 2, . . .,q}* to be that index such that $N(x,y) \cap R_{\Psi}(x,y) \neq \Phi$; *§(x,y,Ri)* is defined as the difference between the testing pixel at *(x,y)* and its adjacent labeled region *Ri*. *§(x,y,Ri)* is calculated as:

$$\delta(x,y,R_i) = \left| g(x,y) - g\left(X_i^c, Y_i^c\right) \right|$$

where, *g(x,y)* indicates the values of the pixel intensity at location *(x,y)* and $g(X^c_i, Y^c_i)$ represents the average values of pixels from homogeneous region *Ri*, with $(X^c_i, Y^c_i)$ as the centroid of *Ri*.
If *N(x,y)* meets two or more of the labeled regions, *Ψ(x,y)* takes a value of *i* such that *N(x,y)* meets *Ri* and *§(x,y,Ri)* is minimized.

$$\varphi(x,y) = \min_{(x,y)\in H} \{\delta(x,y,R_j) | j \in \{1,\ldots,q\}\}$$

Above seeded region growing process is repeated until all pixels in the image have been allocated to the corresponding regions. The above criterions ensure that the final partition of the image is divided into a set of regions as homogeneous as possible on the basic assumptions made initially.

## 4. REGION GROW PROCESS

The goal of region growing is to map the input image data into sets of connected pixels, called regions, according to a prescribed criterion which generally examines the properties of local groups of pixels. The growing starts from a pixel in the proximity of the seed point initially selected by the user. The pixel can be chosen based on either its distance from the seed point or the statistical properties of the neighborhood. Then each of the 4 or 8 neighbours of that pixel are visited to determine if they belong to the same region. This growing expands further by visiting the neighbours of each of these 4 or 8 neighbor pixels. This recursive process continues until either some termination criterion is met or all pixels in the image are examined. The result is a set of connected pixels determined to be located within the region of interest.

Segmentation process becomes semiautomatic starting with an interactive seed point selection step, followed by the region growing process. As a result, the user only needs to find a few representative features and lets the region growing locate all features of similar properties in the same region. This kind of region based method is useful to identify fine blood vessels in MRA and CTA volume data. Researchers developed a pipeline of image processing steps to derive accurate models for visualization and exploration of vascular structures from radiological data [10]. The resulting vessel models are used to study the branching patterns, measurement, etc.

### 4.1 Gradient Based Homogeneity Criteria

Success of region grow algorithm depends on the initial seed selection and criteria used to terminate the recursive region grow process. Hence choosing appropriate criteria is the key in extracting the desired regions. In general, these criteria include region homogeneity, object contrast with respect to background, strength of the region boundary, size, conformity to desired texture features like texture, shape, color.

We used criteria mainly based on region homogeneity and region aggregation using intensity values and their gradient direction and magnitude. This criterion is characterized by a cost function which exploits certain features of images around the seed [11]. These cost function are verified for their match with the specified conditions of homogeneity criteria by comparing their value to be less than 1. If there is a match then pixel under consideration is added to the growing region otherwise excluded from consideration.

Gradient based cost function used in our implementation are defined below

$$\mathrm{Gn} = \sqrt{\mathrm{Gx}^2 + Gy^2} / kG\max$$

such that $0 < Gn < 1$

Where $Gx$ is the horizontal gradient component, $Gy$ is the vertical gradient component; $k$ is the constant parameter which controls the region grow and $Gmax$ is the largest gradient magnitude present in the image.

$Gm = G\max - G(x, y) / G\max - G\min$ such that $0 < Gm < 1$

Where $G(x,y)$ is the gradient magnitude at pixel under consideration and Gmin is the minimum gradient present in the image.

### 4.2 Stack Based Seeded Region Growing Algorithm

We have implemented the 2D seeded region grow algorithm using stack data structure. Since stack is simple to implement and efficient in data access, we used stack to traverse the neighbourhood pixels around the seed location. In our implementation we considered 4-neighbours while growing the region as shown in Figure 2. Similar pseudo code for our implementation is as follows:

```
initialize the stack
    for each seed location
    push seed location to stack
        While (stack not empty)
            pop location
            mark location as Region
            mark location as visited node
  if homogeneity criteria matches for location's left
neighbor pixel
       if left neighbour is not visited
           push left neighbour to stack
if homogeneity criteria matches for location's right
neighbor pixel
       if right neighbour is not visited
           push right neighbour to stack
if homogeneity criteria matches for location's top
neighbor pixel
       if top neighbour is not visited
           push top neighbour to stack
if homogeneity criteria matches for location's
bottom neighbor pixel
       if bottom neighbour is not visited
push bottom neighbor to stack
```

| | | |
|---|---|---|
| x-1, y-1 | **x, y-1** | x+1, y-1 |
| **x-1, y** | **x, y** | **x+1, y** |
| x-1, y+1 | **x, y+1** | x+1, y+1 |

Figure 2- Four Neighbors considered for region grow

Similar concept is extended for segmentation of 3D data set using region grow method. In 3D segmentation, 6 neighbours are considered during segmentation. Two additional pixels along z-axis from 2 adjacent slices are considered along with 4 neighbours shown in Figure 2.

## 5. EXPERIMENTATION

We have created a prototype implementation for the above explained gradient based homogeneity criterion for seeded region grow and tested with medical images. For experimentation purpose we have chosen CT Angiography images of thorax region. In Figure 3 we can see thorax region with ascending and descending aorta. Initially seed is selected on central part of ascending aorta and the region is grown using existing simple SRG method which uses homogeneity criterion based on average pixel density. We can see in Figure 4 that aorta segmentation leading to segmentation of Phrenic nerve appears adjacent to the ascending aorta. When gradient based homogeneity criterion is employed in the algorithm, aorta segmentation is successful by eliminating the spurious adjacent high contrast region getting added to the desired region as shown in Figure 5. By selecting multiple seeds manually at various locations within thorax region, we could segment descending aorta and high contrast bones in the rib cages as shown in Figure 6 to Figure 8.

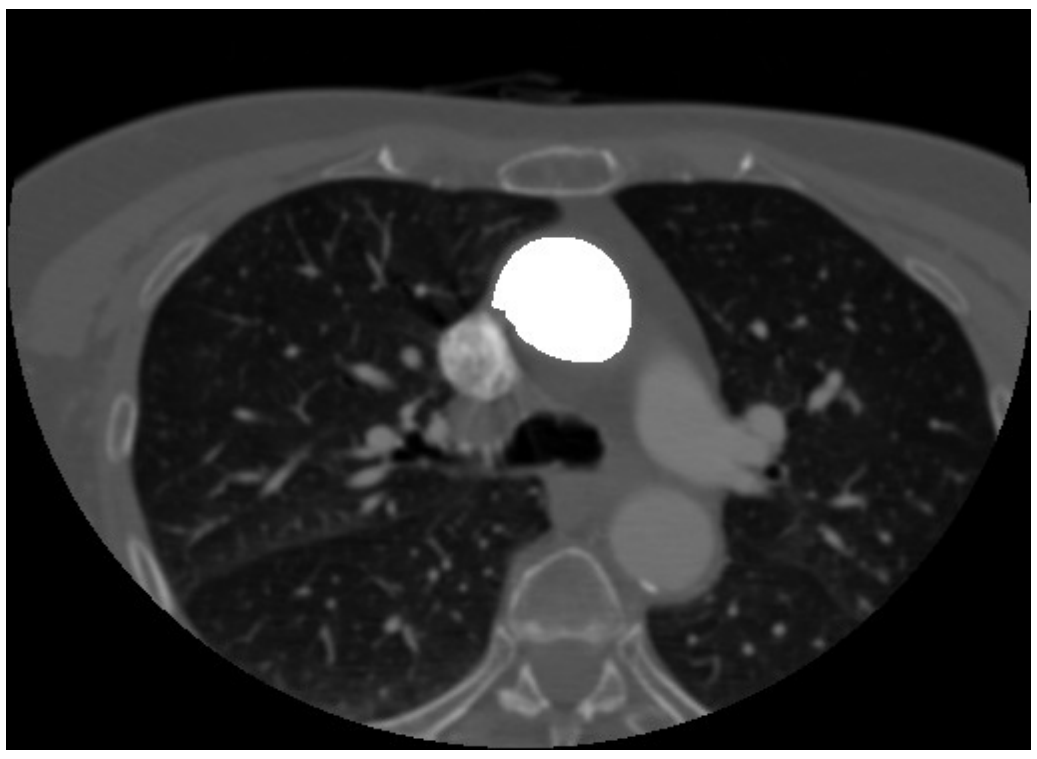

Figure 5- Ascending aorta segmented with Gradient based SRG method with k=0.25

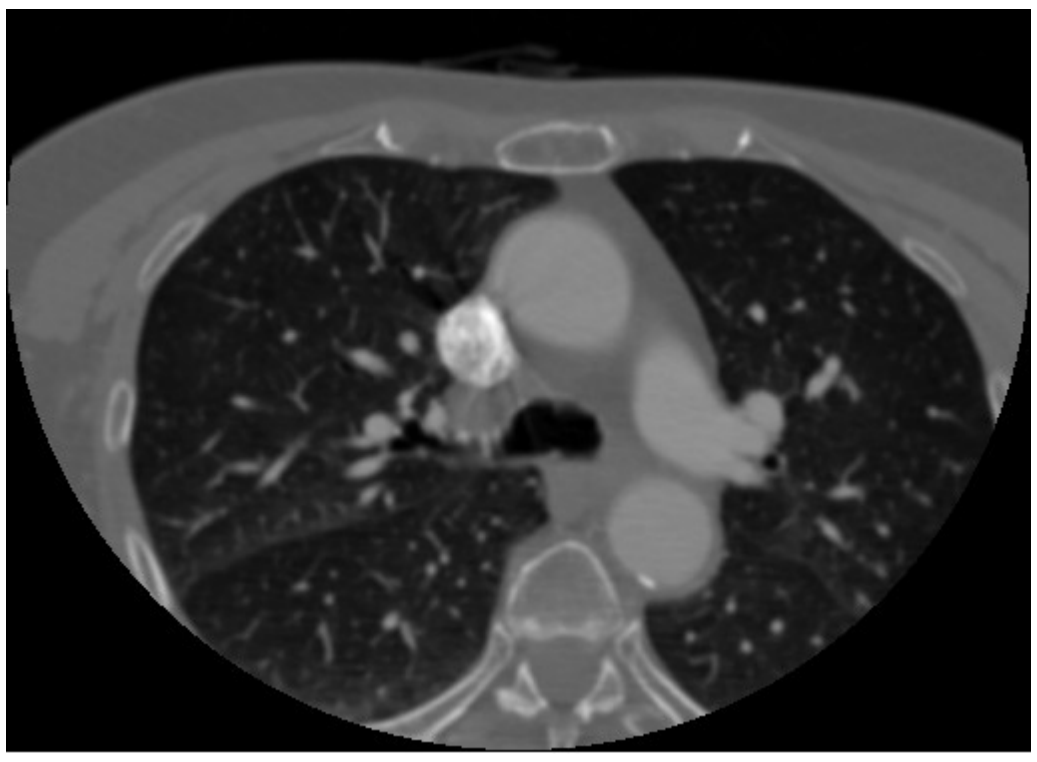

Figure 3- CTA Thorax image

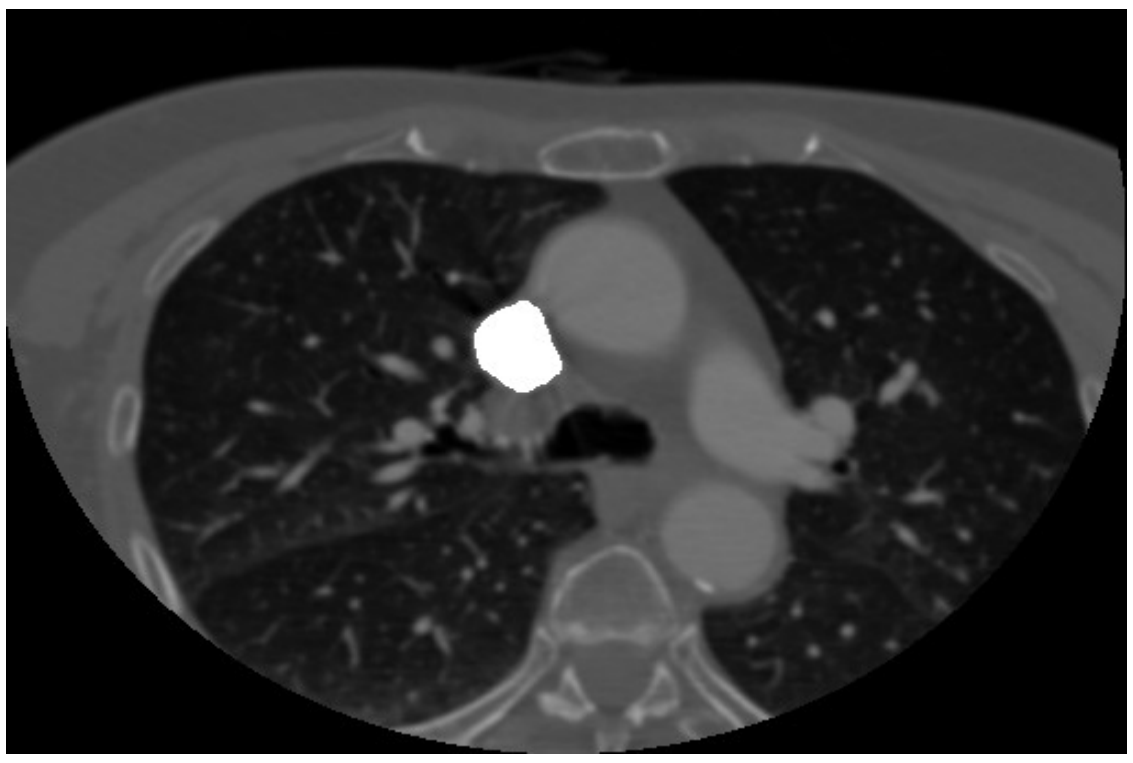

Figure 6- Phrenic nerve segmented with Gradient based SRG method

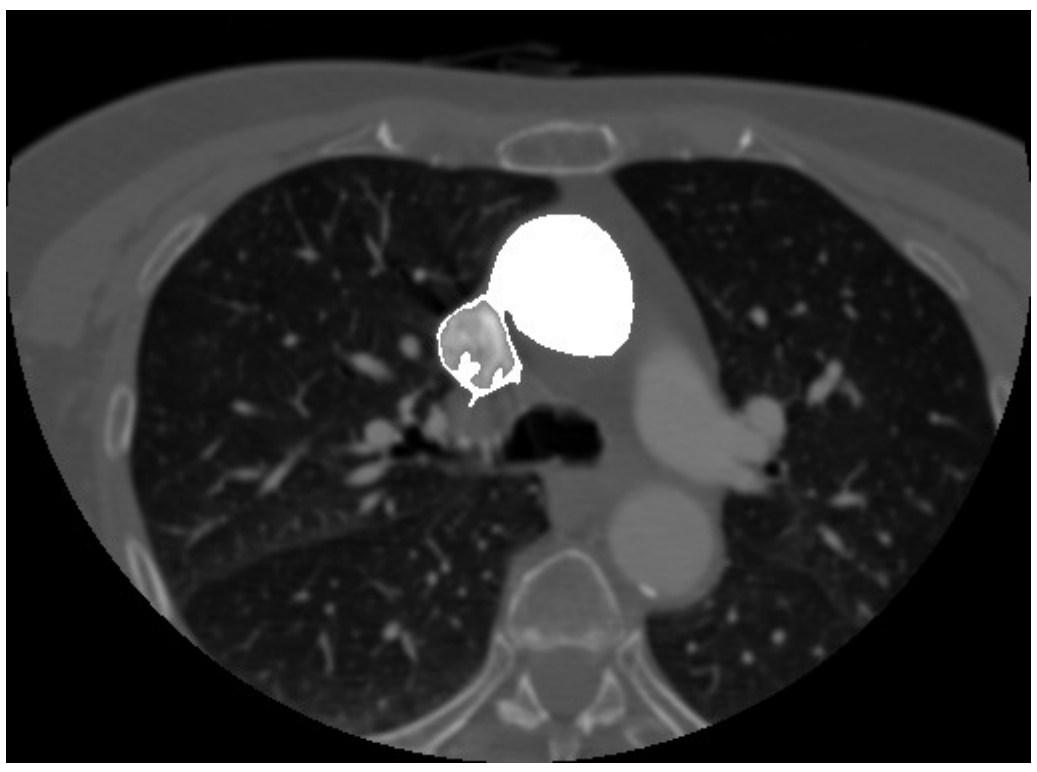

Figure 4- Ascending aorta segmented with Simple SRG method

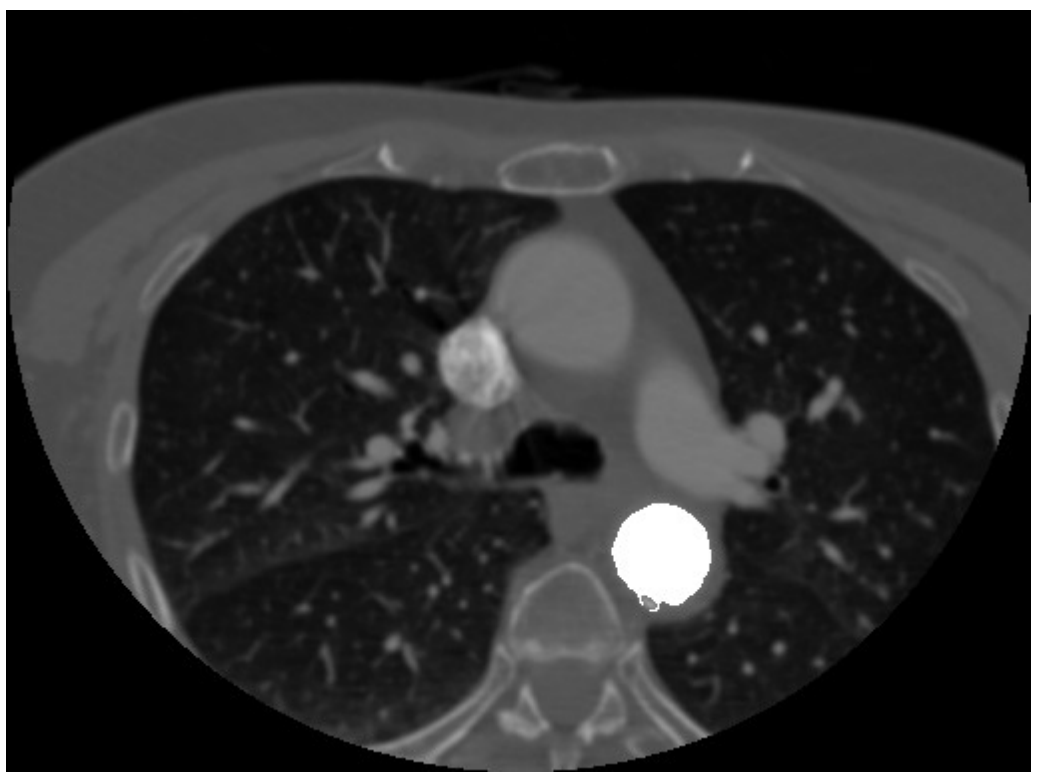

Figure 7- Descending aorta segmented with Gradient based SRG method

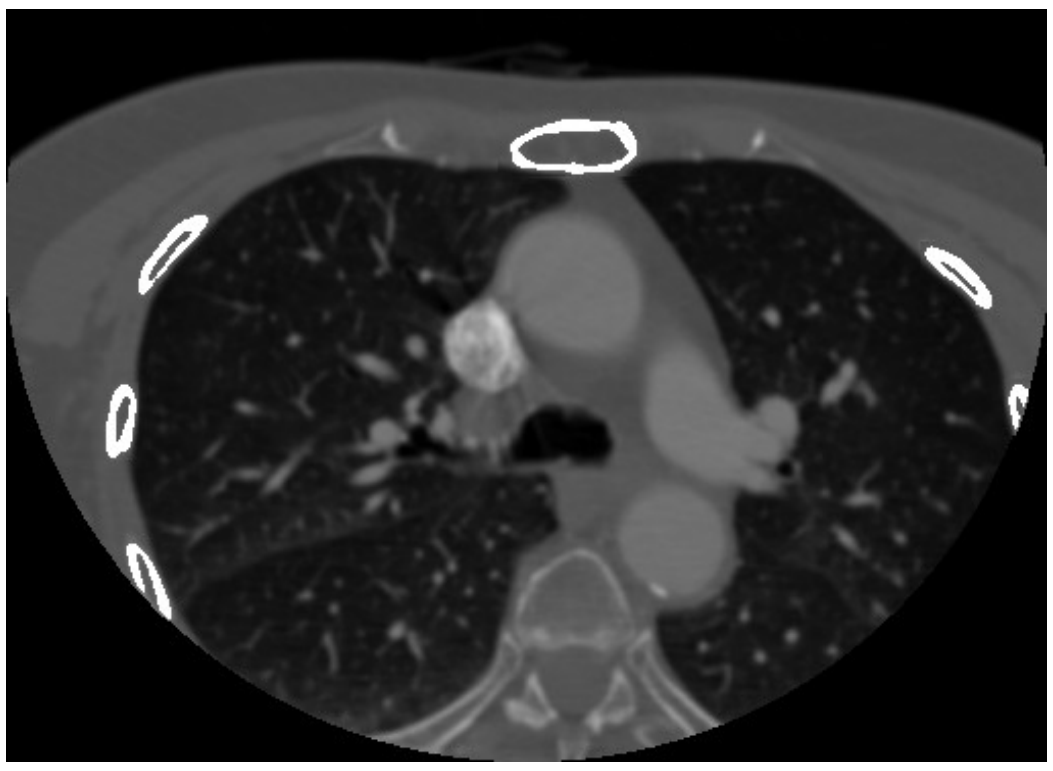

Figure 8- Partial rib bones segmented with Gradient based SRG method

## 6. CONCLUSION

In this study we have presented various aspects of homogeneity criterion selection and its impact on the quality of segmentation in seeded region growing method. Though the region growing through seeding is one of the fundamental segmentation methods, a proper selection of seed location and a selection of homogeneity principle can yield very good results in medical image analysis depending upon the medical case in hand. Here, seed selection is carried out in manual mode and it needs some amount of user intervention in the segmentation process to get better results. Semi-automatic segmentation can be developed where initial seed characteristics can be analyzed to derive subsequent seed positions within the image.

## BIOGRAPHY

**Harikrishna Rai** is a Technical Architect at Software Engineering and Technology Labs of Infosys. His area of interest is Computer Vision and Digital Image Processing with specialization in Medical Imaging. He is having 11 years of experience in research and development at varied industry sectors. He is a BE graduate in Electronics and Communication and MS in Medical Software.

**T.R.Gopalakrishnan Nair** holds M.Tech. (IISc, Bangalore) and Ph.D. degree in Computer Science. He has 3 decades experience in Computer Science and Engineering through research, industry and education. He has published several papers and holds patents in multi domains. He has won the PARAM Award for technology innovation. Currently he is the Director of Research and Industry in Dayananda Sagar Institutions, Bangalore, India.